\documentclass[conference]{IEEEtran}
\IEEEoverridecommandlockouts

\usepackage{amsmath,amssymb}

\usepackage{times}
\usepackage{bm}
\usepackage[numbers]{natbib}
\usepackage{algorithm}
\usepackage{xargs}[2008/03/08]
\usepackage{amssymb}
\usepackage{mathrsfs}
\usepackage{graphicx}
\usepackage{rotating,subfigure}
\usepackage[flushleft]{threeparttable}
\usepackage{multirow}
\usepackage{enumitem} 

\usepackage{caption} 
\usepackage{star}

\usepackage{amsmath,amssymb,amsfonts}
\usepackage{algorithmic}
\usepackage{graphicx}
\usepackage{textcomp}
\usepackage{xcolor}
\def\BibTeX{{\rm B\kern-.05em{\sc i\kern-.025em b}\kern-.08em
		T\kern-.1667em\lower.7ex\hbox{E}\kern-.125emX}}

\begin{document}
	
\title{Understanding the Effects of Second-Order Approximations in Natural Policy Gradient Reinforcement Learning}

\author{\IEEEauthorblockN{Brennan Gebotys}
	\IEEEauthorblockA{
		\textit{University of Waterloo}\\
		bmgebotys@uwaterloo.ca}
	\and
	\IEEEauthorblockN{Alexander Wong}
	\IEEEauthorblockA{
		\textit{University of Waterloo}\\
		a28wong@uwaterloo.ca}
	
	\and
	\IEEEauthorblockN{David A Clausi}
	\IEEEauthorblockA{
		\textit{University of Waterloo}\\
		dclausi@uwaterloo.ca}
}

\maketitle

\begin{abstract}
Natural policy gradient methods are popular reinforcement learning methods that improve the stability of policy gradient methods by utilizing second-order approximations to precondition the gradient with the inverse of the Fisher-information matrix. However, to the best of the authors' knowledge, there has not been a study that has investigated the effects of different second-order approximations in a comprehensive and systematic manner. To address this, five different second-order approximations were studied and compared across multiple key metrics including performance, stability, sample efficiency, and computation time. Furthermore, hyperparameters which aren't typically acknowledged in the literature are studied including the effect of different batch sizes and optimizing the critic network with the natural gradient. Experimental results show that on average, improved second-order approximations achieve the best performance and that using properly tuned hyperparameters can lead to large improvements in performance and sample efficiency ranging up to +181\%. We also make the code in this study available at \url{https://github.com/gebob19/natural-policy-gradient-reinforcement-learning}.
\end{abstract}

\section{Introduction}

The policy gradient method \cite{sutton2000policy} is a popular optimization method for reinforcement learning problems; however, it suffers from unstable training due to high-variance gradient estimates \cite{mohamed2020monte}. A promising solution is to leverage natural policy gradient methods, which preconditions the gradient with the inverse of the Fisher-information matrix \cite{Amari1998} to restrict how much the model can change between training iterations. 

For neural networks, which can have tens of millions of parameters, directly computing the inverse of the Fisher-information matrix is intractable since the Fisher-information matrix is an $n_\theta \times n_\theta$ matrix, where $n_\theta$ is the number of parameters in the model. This led to the research community developing new approximations for the inverse of the Fisher-information matrix which led to having to choose the most appropriate approximation method. Furthermore, to generate effective performance, setting the batch size and how to optimize the critic network are also important details. To the best of the authors’ knowledge, there has not been a study that has holistically investigated strategies for choosing these details to achieve high performance with natural policy gradient methods in a comprehensive and systematic manner.

In this paper, we study the effects of five different second-order approximations: Hessian-free optimization (HF), diagonal approximations (Diagonal), Kronecker-factored approximate curvature (KFAC), Eigenvalue-corrected Kronecker factorization approximate curvature (EKFAC), and Time-efficient natural gradient descent (TENGraD). Across these approximations, we investigate hyperparameters which aren't typically given importance in policy gradient research including, the effect of using different batch sizes and the effect of using different optimization methods on the critic network. We also study these approximations and hyperparameters across multiple reinforcement learning environments and record statistics on performance, training stability, sample efficiency, and computation time. 

\section{Background}

In this section, we first define the notation used throughout the paper, followed by a brief overview of the policy gradient method, the natural policy gradient method, and the five different second-order approximations which can be used to implement the natural policy gradient method. 

\newcommand{\rew}{r}
\newcommand{\E}{\mathbb{E}}
\newcommand{\thetaK}[1]{\theta^{(#1)}}
\newcommand{\DKL}{D_{KL}}
\newcommand{\inv}[1]{{#1}^{-1}}

We consider an infinite-horizon discounted Markov decision process, defined by the tuple ($\mathcal{S}$, $\mathcal{A}$, $p$, $\rew$, $\gamma$) which defines the state distribution $\mathcal{S}$, action distribution $\mathcal{A}$, state transition function $p$, reward function $\rew$, and discount factor $\gamma$. Specifically, at time $t$, an agent is in a state $s_t \in \mathcal{S}$ and samples an action $a_t \in \mathcal{A}$ from its policy, $\pi (a_t | s_t)$. The environment then produces a reward $\rew(a_t, s_t)$ and a new state $s_{t+1} \in \mathcal{S}$ according to the transition probability function $p(s_{t+1} | a_t, s_t)$. Repeating this process $n$ times produces a trajectory, $\tau$, of length $n$.

\subsection{Policy Gradient Methods}

The policy gradient method \cite{sutton2000policy} parameterizes its policy with parameters $\theta$ and aims to minimize the negative expected reward of its trajectories:

\begin{equation}
    \underset{\theta}{\argmin} \, L(\theta) = \underset{\theta}{\argmin} \, - \E_{\tau \sim \pi(\tau | \theta)} [\rew(\tau) \,|\, \pi(\cdot | \theta)]
    \label{eqloss}
\end{equation}

\noindent where for each time step $t$ in the trajectory $\tau$: $\rew(\tau)_t = \sum_{i=0}^\infty \gamma^i \rew(a_{t+i}, s_{t+i})$ is the discounted cumulative reward after taking action $a_t$ in state $s_t$ and $\pi(\cdot | \theta)_t = \pi (a_t | s_t, \theta)$ is the policy's probability of taking action $a_t$. In general, to optimize \eqref{eqloss}, the policy gradient method uses the following gradient:

\begin{align}
\nabla_\theta L(\theta) &= - \E_{\tau \sim \pi(\tau | \theta)} [\Psi \nabla_\theta \text{log} \pi(\cdot | \theta)] \label{pgrad}
\end{align}

\noindent where $\Psi$ is typically chosen as an advantage function, $A$, which has the general form of: $A(a_t, s_t) = \rew(\tau)_t - V(s_t)$, where $V(s_t)$ is the expected cumulative reward of the policy in state $s_t$. In our experiments, we set $\Psi$ to use the generalized advantage estimation method (GAE) \cite{gae} which involves training a critic network to estimate $V(s_t)$. While the policy gradient method directly optimizes \eqref{pgrad}, doing so often results in high-variance gradient estimates which leads to unstable training and poor performance \cite{mohamed2020monte}.

\subsection{Natural Policy Gradient Methods}

To stabilize training, the natural policy gradient method  \cite{Amari1998} restricts how much the policy can change across training iterations by adding a Kullback–Leibler divergence constraint ($\DKL$) between policy iterations:

\begin{equation}
    \thetaK{k+1} = \underset{\thetaK{k+1}}{\argmin} \, L(\thetaK{k+1}) + \lambda \DKL(\pi(\cdot | \thetaK{k}), \pi(\cdot | \thetaK{k+1})) \label{ngrad_deriv}
\end{equation}

\noindent where $\thetaK{k}$ defines the parameters at optimization iteration $k$. To approximately solve \eqref{ngrad_deriv}, we use a first-order taylor-series approximation of $L$ around $\thetaK{k}$, a second-order taylor-series approximation of $\DKL$, and solve for the optimum (see \cite{natural_lec} and \cite{natural_grad_blog} for a full derivation). In doing so, we derive the natural policy gradient method, 

\begin{equation}
    \thetaK{k+1} = \thetaK{k} - \inv{\lambda} \inv{F} \nabla L(\thetaK{k})
    \label{ngrad_def}
\end{equation}

\noindent where $F$ is the Fisher-information matrix: 

\begin{equation}
    F = \E_{\pi(\cdot |\theta)} [\nabla_\theta \log \pi(\cdot |\theta) \nabla_\theta \log \pi(\cdot |\theta)^T]
\end{equation}

\begin{figure*}[th]
\vskip 0.2in
\begin{center}
\includegraphics[width=\textwidth]{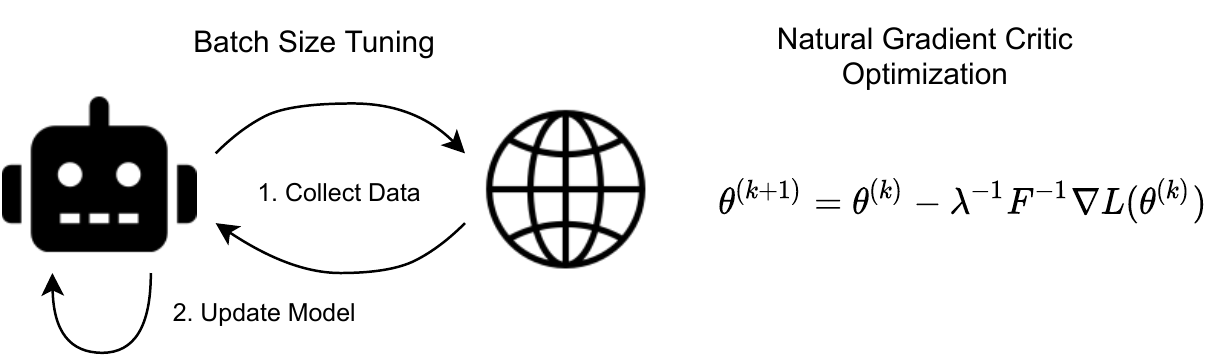}
\caption{Diagram of the two hyperparameters introduced in this paper for natural policy gradient methods: tuning the batch size for performance and optimizing the critic method using the natural gradient. These strategies can be applied across all second-order approximations for any reinforcement learning environment.}
\label{fig:strats}
\end{center}
\vskip -0.2in
\end{figure*}

\section{Second-Order Approximations}

A key problem is that the Fisher-information matrix ($F$) is an $n_\theta \times n_\theta$ matrix, where $n_\theta$ is the number of parameters in the model. For neural networks which can have tens of millions of parameters, directly computing the inverse of the Fisher-information matrix is intractable. This has led to rising interest in the research community to develop new ways to approximate the inverse of the Fisher-information matrix for neural networks. However, most of these approximations have never been fully investigated in reinforcement learning (RL) tasks. Below, we review five of the most popular approximations and their related work in RL.

\vspace{.5em}

\textbf{Diagonal Approximations (Diagonal)} \cite{diagonal}: Diagonal approximations use a diagonal approximation of $F$. This makes storage and inversion easy, however, it ignores a large amount of curvature information. While some RL research \cite{gu2017qprop} indirectly use the diagonal approximation by using the Adam optimizer \cite{kingma2017adam}, there has not been any specific focus on understanding the fundamental trade-offs when using Diagonal approximations compared to other approximations when realizing the natural policy gradient.

\textbf{Hessian-free Approximations (HF)} \cite{hessianfree}: Hessian-free approximations use matrix-vector products and the conjugate gradient method to approximately solve for $\inv{F} \nabla L(\theta)$. However, computing the conjugate gradient at each iteration can be computationally expensive. One of the most popular implementations of the natural policy gradient which used the HF approximation is Trust region policy optimization (TRPO) \cite{trpo} which used a backtracking line search method to achieve state-of-the-art results in continuous control tasks at the time.

\textbf{Kronecker-factor Approximations (KFAC/EKFAC)} \cite{kfac}: Kronecker-factor approximations use a block-diagonal approximation of $F$ which assumes each layer is independent of the others. KFAC \cite{kfac} decomposes the Fisher-information matrix using the Kronecker product. EKFAC \cite{ekfac} uses a different approximation based on KFAC which showed improved results across computer vision experiments. In RL, Actor-critic using kronecker-factored trust region (ACKTR) \cite{acktr} used KFAC optimization and step-size clipping to realize the natural policy gradient and showed improved results compared to TRPO. EKFAC has never been investigated in the RL literature.  

\textbf{Woodbury Approximations (TENGraD)} \cite{tengrad}: While EKFAC proposes an improved approximation to KFAC, TENGraD provides an exact block-diagonal approximation using the Woodbury identity. However, TENGraD requires constructing a $m \times m$ matrix where $m$ is the batch size which can be computationally expensive when using large batch sizes. TENGraD has never been investigated in the RL literature.

\section{Key Hyperparameters}

In this section, we give a brief introduction to the hyperparameters we investigate. This includes batch size tuning and using the natural gradient to optimize the critic network. Figure \ref{fig:strats} shows a diagram of the two hyperparameters. 

\subsection{Batch Size Tuning}

In the policy gradient setting, the batch size defines how many training examples to collect in the environment before using the collected data to update the model. Using a larger batch size will result in more examples, for a more stable update, but fewer model updates per environment step. Comparatively, using a smaller batch size will result in fewer examples, for a less stable update, but more updates per environment step. Although the importance of the batch size is not emphasized much in the literature, we find that tuning the batch size is a critical hyperparameter that can significantly affect the final performance and other metrics of the model (a similar result was found for Q-learning methods \cite{Kielak2020}). 

To understand how a tuned batch size can improve the baseline performance, we tune the batch size of the baseline models based on the performance metric.

\begin{figure*}[th]
\begin{center}
\includegraphics[width=\textwidth]{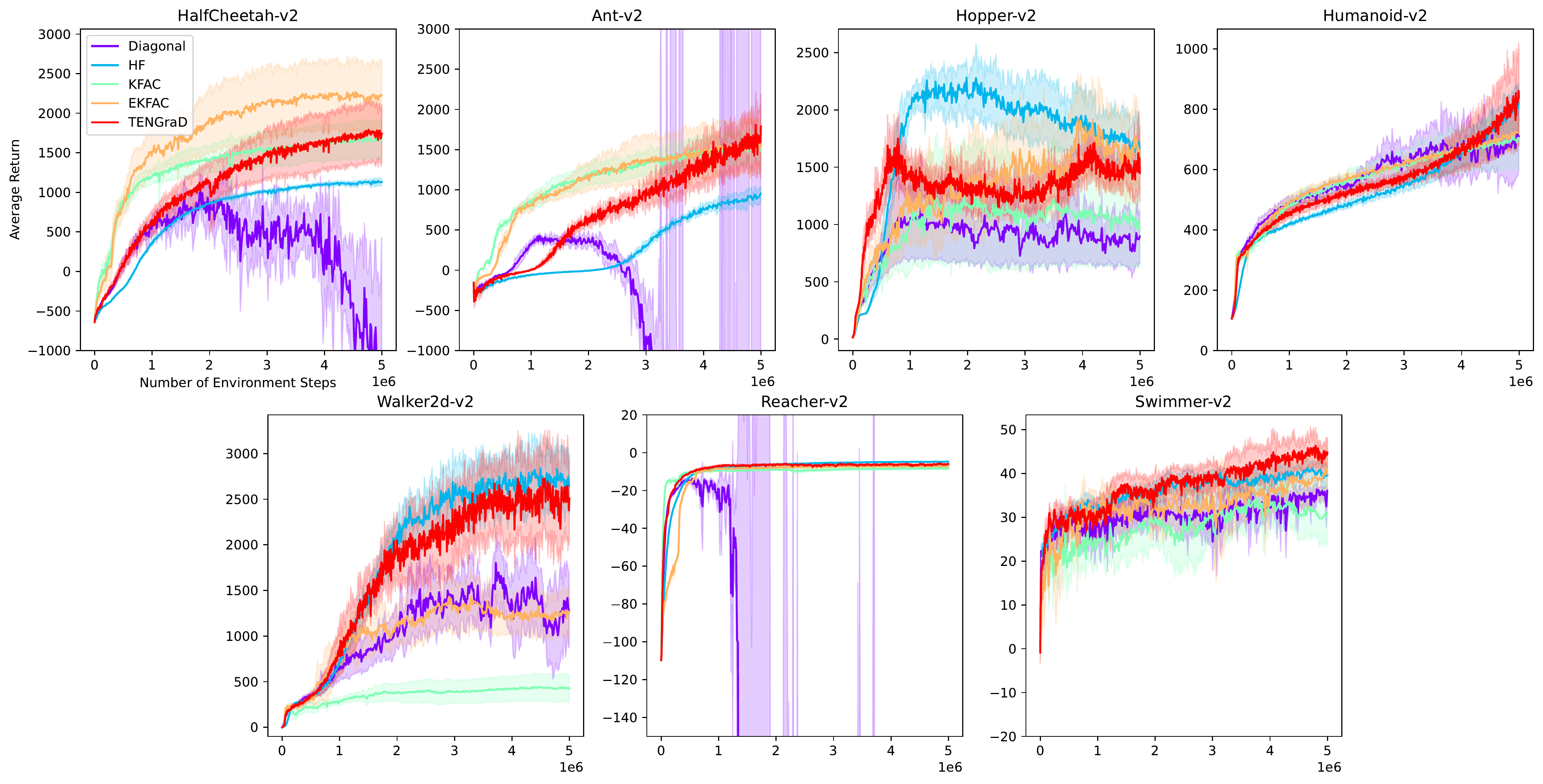}
\caption{Average return vs the number of environment steps of five different second-order approximations across seven different MuJoCo environments using the baseline hyperparameters. All experiments are across 10 random seeds. The solid-colored line represents the mean value across the different seeds and the shaded region represents one standard deviation above and below the mean. TENGraD (red) achieves the best performance across most environments. HF is the most stable across seeds.}
\label{fig:baseline}
\end{center}
\end{figure*}

\begin{table*}[ht]
    \centering
    \begin{tabular}{lrrrr}
    \toprule
    {Approximation} &  Performance &  Stability &  Sample Efficiency &   Computation Time \\
    \midrule
    Diagonal &        12.52 &   -12338.0 &            -2170.0 & \textbf{-37.051} \\
    HF       &       68.73 &     \textbf{-107.0} &            -1708.0 & -37.282 \\
    KFAC     &        29.03 &     -178.0 &            \textbf{-1263.0} & -37.430 \\
    EKFAC    &       61.85 &     -235.0 &            -1316.0 & -37.624 \\
    TENGraD  &       \textbf{85.69} &     -182.0 &            -1425.0 & -52.861 \\
    \bottomrule
    \end{tabular}
    \caption{Mean metrics of five different approximations across seven different MuJoCo environments. The best result for each metric is in bold. Larger values are better. TENGraD achieves the best performance score; HF achieves the best stability score; KFAC achieves the best sample efficiency score, and Diagonal achieves the best computation time score.}
    
    \label{tab:baseline}

\end{table*}

\subsection{Natural Critic Optimization}

Another hyperparameter is to improve the optimization of the critic network using the natural gradient. While some work investigated using clipped targets to stabilize training \cite{engstrom2020implementation}, ACKTR \cite{acktr} used the natural gradient to further improve the stability and showed improved performance using KFAC optimization for both the policy and the critic network. 

While ACKTR investigated applying only KFAC optimization to both the policy network and the critic network, we investigate five different second-order approximations for the policy network and for each of these we investigate five second-order approximations for tuning the critic network. To achieve this, we leverage the models with a tuned batch size and use grid search to find the best approximation to optimize the critic network with. The results are shown in Section \ref{sec:training_strats}.

\section{Experiments}

In this section, we investigate the performance of the five second-order approximations and the effect of our hyperparameters. Furthermore, in doing so, we also investigate the performance of EKFAC and TENGraD approximations for natural policy gradient reinforcement learning for the first time in the literature. We compare each approximation across seven different MuJoCo environments: HalfCheetah-v2, Ant-v2, Hopper-v2, Humanoid-v2, Walker2d-v2, Reacher-v2, and Swimmer-v2. All experiments were run across 10 different random seeds. 

Throughout the experiments, we study four metrics: \textbf{Performance} which is the maximum average return achieved by the policy across training; \textbf{Stability} which is the average standard deviation (across seeds) across training; \textbf{Sample Efficiency} which is the number of environment steps to achieve a specific performance threshold; and \textbf{Computation Time} which is the amount of time to take 100K environment steps while training. More information on these metrics is available in Appendix \ref{sec:metrics}. 

For clarity purposes, to make all metrics correspond with `larger is better', we flip the sign of the metrics where smaller values are better (i.e., Stability, Sample Efficiency, and Computation Time). For example, if two agents achieve an average standard deviation (across seeds) of 500 and 600 (where 500 is a smaller standard deviation and is thus more stable), we flip the signs to get -500 and -600, in which case the policy with the larger value of -500 is more stable and is therefore considered better. All hyperparameter information can be found in Appendix \ref{sec:hparams}.

\subsection{Baseline}

\begin{figure*}[h]
\begin{center}
\includegraphics[width=\textwidth]{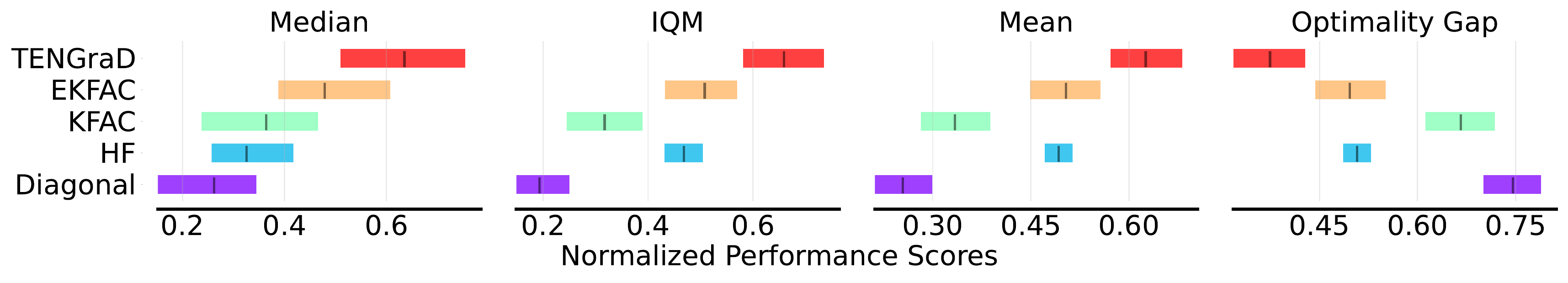}
\caption{Normalized performance scores of all second-order approximations across four aggregate metrics with 95\% stratified bootstrap confidence intervals. A larger score in the median, IQM, and mean aggregations and a smaller score in the optimality gap aggregation corresponds to better performance. We see TENGraD statistically outperforms all other approximations across IQM, mean, and the optimality gap aggregation metrics. The median aggregate metric shows a trend that newer second-order approximations lead to better median performance.}
\label{fig:metrics_baseline}
\end{center}
\end{figure*}

\begin{table*}[htp]
    \centering

    \begin{tabular}{lllll}
    \toprule
    {Environment} &  Performance &     Stability & Sample Efficiency &       Speed \\
    \midrule
    HalfCheetah-v2 &  1687 (+67\%) &   -439 (+16\%) &      -2775 (-50\%) &  -22 (+18\%) \\
    Ant-v2         &   \textbf{825 (+94\%)} &  \textbf{-178 (+100\%)} &      \textbf{-1025 (+10\%)} &  -43 (+18\%) \\
    Hopper-v2      &   527 (-55\%) &    -48 (+87\%) &               NaN &  -27 (+16\%) \\
    Humanoid-v2    &   486 (-33\%) &    -20 (+72\%) &               NaN &  \textbf{-50 (+19\%)} \\
    Walker2d-v2    &   526 (-71\%) &    -59 (+83\%) &     -2175 (-222\%) &  -28 (+18\%) \\
    Reacher-v2     &   -12 (+14\%) &   -304 (+92\%) &       -625 (-34\%) &  \textbf{-21 (+19\%)} \\
    Swimmer-v2     &    33 (-10\%) &      -4 (+7\%) &               NaN &  -22 (+15\%) \\
    \bottomrule
    \end{tabular}
    \caption{Performance of the \textit{Diagonal} approximation across seven different environments with a tuned batch size and where the critic network is optimized with the natural gradient. The percentage improvement compared to the baseline is in brackets. The largest improvement for each metric is in bold. We see performance, stability and sample efficiency improvements up to +94\%, +100\%, and +10\% with up to a +19\% improvement in the speed score.}
    \label{tab:diagonal_tuned}

\end{table*}
    
\begin{table*}[htp]
    \centering

    \begin{tabular}{lllll}
    \toprule
    {Environment} &  Performance &     Stability & Sample Efficiency &       Speed \\
    \midrule
    HalfCheetah-v2 &  1506 (+32\%) &  -216 (-423\%) &      \textbf{-1120 (+62\%)} &   -29 (-6\%) \\
    Ant-v2         &  \textbf{1573 (+65\%)} &  -143 (-189\%) &      -1370 (+57\%) &   -54 (-3\%) \\
    Hopper-v2      &   504 (-78\%) &    -45 (+85\%) &               NaN &   -34 (-3\%) \\
    Humanoid-v2    &   499 (-39\%) &    -13 (+60\%) &               NaN &   -62 (-3\%) \\
    Walker2d-v2    &   398 (-86\%) &    \textbf{-35 (+89\%)} &               NaN &   -36 (-4\%) \\
    Reacher-v2     &    -4 (+11\%) &     -0 (+63\%) &       -270 (+44\%) &   -29 (-9\%) \\
    Swimmer-v2     &    47 (+15\%) &     -5 (-57\%) &       -820 (+35\%) &  -30 (-12\%) \\
    \bottomrule
    \end{tabular}
    \caption{Performance of the \textit{HF} approximation across seven different environments with a tuned batch size and where the critic network is optimized with the natural gradient. The percentage improvement compared to the baseline is in brackets. The largest improvement for each metric is in bold. We see performance, stability and sample efficiency improvements up to +65\%, +89\%, and +62\% with at most a 12\% decrease in the speed score.}
    \label{tab:hf_tuned}

\end{table*}
    
\begin{table*}[htp]
    \centering

    \begin{tabular}{lllll}
    \toprule
    {Environment} &  Performance &     Stability & Sample Efficiency &      Speed \\
    \midrule
    HalfCheetah-v2 &  2319 (+38\%) &   -396 (-52\%) &       -480 (+29\%) &  -27 (-1\%) \\
    Ant-v2         &  \textbf{2593 (+65\%)} &  -386 (-106\%) &       -240 (+38\%) &  -52 (+0\%) \\
    Hopper-v2      &   494 (-60\%) &    \textbf{-86 (+85\%)} &               NaN &  \textbf{-32 (+2\%)} \\
    Humanoid-v2    &   451 (-36\%) &    -34 (+48\%) &               NaN &  -62 (-0\%) \\
    Walker2d-v2    &   337 (-24\%) &    -95 (+36\%) &               NaN &  -35 (+1\%) \\
    Reacher-v2     &    -7 (+19\%) &     -1 (-40\%) &       -360 (-14\%) &  -26 (+1\%) \\
    Swimmer-v2     &    52 (+53\%) &    -15 (-87\%) &     \textbf{ -1770 (+53\%)} &  -27 (+1\%) \\
    \bottomrule
    \end{tabular}
    \caption{Performance of the \textit{KFAC} approximation across seven different environments with a tuned batch size and where the critic network is optimized with the natural gradient. The percentage improvement compared to the baseline is in brackets. The largest improvement for each metric is in bold. We see performance, stability and sample efficiency improvements up to +65\%, +85\%, and +53\% with at most a 1\% decrease in the speed score.}
    \label{tab:kfac_tuned}

\end{table*}
    
\begin{table*}[ht]
    \centering

    \begin{tabular}{lllll}
    \toprule
    {Environment} &  Performance &    Stability & Sample Efficiency &      Speed \\
    \midrule
    HalfCheetah-v2 &  2513 (+11\%) &  -603 (-17\%) &       -570 (+0\%) &  -27 (-1\%) \\
    Ant-v2         &  \textbf{2345 (+50\%)} &  -325 (-10\%) &      \textbf{ -375 (+38\%)} &  -52 (-0\%) \\
    Hopper-v2      &   495 (-74\%) &  \textbf{ -66 (+86\%)} &               NaN &  \textbf{-33 (+4\%)} \\
    Humanoid-v2    &   456 (-37\%) &   -21 (+35\%) &               NaN &  -63 (-0\%) \\
    Walker2d-v2    &   403 (-72\%) &   -72 (+77\%) &               NaN &  -36 (-2\%) \\
    Reacher-v2     &    -6 (+10\%) &    -1 (+36\%) &        -615 (-5\%) &  -27 (-2\%) \\
    Swimmer-v2     &    37 (-11\%) &    -7 (+12\%) &     -2910 (-169\%) &  -27 (-1\%) \\
    \bottomrule
    \end{tabular}
    \caption{Performance of the \textit{EKFAC} approximation across seven different environments with a tuned batch size and where the critic network is optimized with the natural gradient. The percentage improvement compared to the baseline is in brackets. The largest improvement for each metric is in bold. We see EKFAC achieves improvements in performance, stability, and sample efficiency of up to +50\%, +86\%, and +38\% respectively with at most a 1\% decrease in the speed score.}
    \label{tab:ekfac_tuned}

\end{table*}

\begin{table*}[ht]
    \centering

    \begin{tabular}{lllll}
    \toprule
    {Environment} &   Performance &     Stability & Sample Efficiency &      Speed \\
    \midrule
    HalfCheetah-v2 &   2058 (+15\%) &   -592 (-69\%) &       -788 (+52\%) &  -44 (-5\%) \\
    Ant-v2         &  3855 (+113\%) &  -464 (-153\%) &       -652 (+59\%) &  -71 (-4\%) \\
    Hopper-v2      &   1510 (-14\%) &   -148 (+35\%) &    -4256 (-1008\%) &  -50 (-4\%) \\
    Humanoid-v2    &    640 (-26\%) &   \textbf{ -21 (+60\%)} &               NaN &  -79 (-2\%) \\
    Walker2d-v2    &   1614 (-41\%) &   -267 (+41\%) &      -1296 (-89\%) &  -53 (-4\%) \\
    Reacher-v2     &     -5 (+13\%) &     -1 (-16\%) &       -208 (+43\%) &  -43 (-4\%) \\
    Swimmer-v2     &   \textbf{130 (+181\%)} &   -49 (-874\%) &       \textbf{-172 (+86\%)} &  -44 (-4\%) \\
    \bottomrule
    \end{tabular}
    \caption{Performance of the \textit{TENGraD} approximation across seven different environments with a tuned batch size and where the critic network is optimized with the natural gradient. The percentage improvement compared to the baseline is in brackets. The largest improvement for each metric is in bold. We see TENGraD achieves improvements in performance, stability, and sample efficiency of up to +181\%, +60\%, and +86\% respectively with at most a 5\% decrease in the speed score.}
    \label{tab:tengrad_tuned}

\end{table*}

We first investigate each approximation's baseline performance using the same batch size and using SGD to optimize the critic network. The training curves of the different approximations are shown in Figure \ref{fig:baseline}. The solid-colored line represents the mean value across the random seeds and the shaded region represents one standard deviation above and below the mean. We also show the mean metrics across the seven different environments in Table \ref{tab:baseline}.

\textbf{Performance} Since each environment has different reward scales, to weigh each environment's score equally, we normalize the performance scores in Table \ref{tab:baseline} and Figure \ref{fig:metrics_baseline}. 

Figure \ref{fig:baseline} shows that no approximation outperforms all other approximations across all environments and shows that the best approximation depends on the environment. Table \ref{tab:baseline} shows TENGraD achieves the best performance score across all approximations, followed by HF and EKFAC, followed by KFAC and lastly Diagonal. Table \ref{tab:baseline} also shows that improved approximations can lead to improved performance since TENGraD achieves the best mean performance score across all environments and is the most accurate approximation. We also see that the Diagonal achieves the lowest performance score and that in some environments (Ant-v2 and Reacher-v2 shown in Figure \ref{fig:baseline}) it diverges near the end of the training; likely because it ignores a large amount of curvature leading to unstable updates. 

Figure \ref{fig:metrics_baseline} shows normalized performance scores across four aggregate metrics with 95\% stratified bootstrap confidence intervals from  \cite{agarwal2021deep} including median, interquartile mean (IQM), mean, and the optimality gap across all approximations. A larger score in the median, IQM, and mean aggregations and a smaller score in the optimality gap aggregation corresponds to better performance. 

We see TENGraD significantly outperforms the other approximations in terms of the IQM, mean, and optimality gap aggregation metrics because it achieves the best score and there is no overlap between its confidence intervals and the others. EKFAC and HF achieve the second-best scores however, their confidence intervals overlap which shows their performance is not significantly different from each other. Lastly, we see KFAC achieves the third-best score, followed by Diagonal with the worst score. Analyzing the median metrics in Figure \ref{fig:metrics_baseline} is not as straightforward due to a large amount of overlap in confidence intervals; this is not surprising since the median is a poor measure of overall performance \cite{agarwal2021deep}. We see TENGraD is significantly better than all other approximations other than EKFAC; and that EKFAC is significantly better than Diagonal. Interestingly, the median aggregate metric shows a trend that newer second-order approximations lead to better median performance.

\textbf{Stability} HF achieves the best stability score, closely followed by KFAC, followed by TENGraD and EKFAC, and lastly, with a significantly lower score, Diagonal. HF achieves the best stability score, likely because it utilizes the conjugate gradient algorithm for its update. In contrast, Diagonal achieves the lowest score, likely because its approximation ignores a large amount of curvature leading to inaccurate gradient scaling. We see in the HalfCheetah-v2 and Reacher-v2 environments, Diagonal is unable to remain stable throughout training and diverges at the end of training. 

\textbf{Sample Efficiency} KFAC achieves the best sample efficiency score, closely followed by EKFAC and TENGraD, followed by HF and Diagonal.

\textbf{Computation Time} Diagonal achieves the best computation time score, followed by HF, followed by KFAC and EKFAC, and lastly TENGraD. Diagonal achieves the best scores due to its simple and efficient implementation, only requiring an element-wise multiplication and division. Whereas, TENGraD achieves the lowest score due to it having to periodically compute the inverse of a very large $m \times m$ matrix where $m$ is the batch size. 

\subsection{Hyperparameters} \label{sec:training_strats}

We now investigate how we can leverage a performance-tuned batch size and use the natural gradient to optimize the critic network to achieve performance improvements across all approximations in Tables \ref{tab:diagonal_tuned} - \ref{tab:tengrad_tuned}. For succinctness, we move the results of batch size tuning in Appendix \ref{sec:batch_results}. In all the tables we show the percentage change compared to the baseline model in brackets and bold the largest improvement for each metric. If the policy is unable to achieve the sample efficiency threshold compared to the baseline we show `NaN'. 

\textbf{Key Trends:} Overall, we see tuning the batch size and using the natural gradient to optimize the critic network can lead to significant improvements in performance, training stability, sample efficiency, and speed in many reinforcement learning environments across all five natural gradient approximations. Specifically, we see TENGraD achieves improvements in performance, stability, and sample efficiency of up to +181\%, +60\%, and +86\% respectively with at most a 5\% decrease in the speed score. The results of applying the strategies across all the other approximations are summarized in their respective table captions. 

Interestingly, we see that using the hyperparameters the metrics are usually improved in the HalfCheetah-v2, Ant-v2, Reacher-v2, and Swimmer-v2 environments, however, the other environments such as Hopper-v2, Humanoid-v2, and Walker2d-v2 see results decrease compared to the baseline. This may be because the strategies are tuned on the HalfCheetah-v2 environment which is much similar to the environments in which the policy performed well in and is not similar to the environments in which the policy performed poorly. We hypothesise that tuning the strategies on the poor performance environments can result in similar performance improvements seen in the other environments. We leave this for future research.

Another trend we see is that when the performance score is improved usually the stability score decreases; and when the performance score is improved, we see a similar improvement in the sample efficiency score. 

\section{Conclusion}

In this paper, we studied five different second-order approximations across multiple key metrics to better understand how each approximation affects the performance of the natural policy gradient. Furthermore, we investigated the effect of two different hyperparameters: the batch size and the optimization method for the critic network, and showed they have a large effect on the final performance, even though they aren't typically acknowledged in the literature.

We found that properly tuning the hyperparameters can lead to large improvements in performance and sample efficiency ranging up to +181\% and +86\% respectively across the MuJoCo control benchmarks and that TENGraD achieved the best performance out of all the approximations. 

We also find that there is a fundamental trade-off between achieving high performance and maintaining stability throughout training, and that performance is positively correlated with sample efficiency. We hope this research helps expand our field's understanding of how different approximations and hyperparameters affects the natural policy gradient and helps practitioners efficiently leverage natural policy gradient methods for real-world reinforcement learning tasks. 

\bibliography{main}
\bibliographystyle{abbrv}

\newpage
\appendix

\section{Tuned Batch Size Results} \label{sec:batch_results}

We show the results of tuning the batch size across all approximations in Tables \ref{tab:diag_batch_tuned} - \ref{tab:tengrad_batch_tuned}. We found KFAC's tuned batch size was the same as its baseline batch size so we don't include its results. We see large improvements in performance, stability, sample efficiency, and speed across all approximations showing that tuning the batch size is a key strategy for achieving the best performance.

\section{More Details on the Metrics} \label{sec:metrics}

For the sample efficiency metric, the performance thresholds are computed for each environment as the lowest performance score across all approximations in the baseline experiments. The thresholds are then held constant throughout the rest of the experiments. For clarity purposes, the threshold scores are divided by 1000. The exact thresholds used are made available in our code repository.

In Table \ref{tab:baseline}, since each environment has different reward scales, to weigh each environment's score equally, we normalize the performance scores of each environment by dividing by the highest score achieved in that environment across all approximations and multiplying it by 100, making the largest achievable reward 100. We don't scale the values by 100 for Figure \ref{fig:metrics_baseline}.

\section{Hyperparameters} \label{sec:hparams}

\subsection{Baseline}
For the baseline model, we tuned the following hyperparameters for one million environment steps: damping parameter with values (1e-1 1e-2), critic learning rate with values  (1e-2 1e-3), whether to use a backtracking line search or not based on \cite{trpo}, when not using line search we use step-size clipping used in \cite{acktr} and tune the maximum learning rate with values (1e-1 1e-2). All approximations used a batch size of 15K except for TENGraD which used a batch size of 6K because our GPUs didn't have enough memory for a batch size of 15K. We chose 15K because it is commonly used in the literature \cite{trpo, acktr}.

Although TENGraD uses a smaller batch size which may be seen as unfair due to the increased number of optimization steps, comparing all approximations with their tuned batch sizes (with available batch sizes smaller than TENGraD's batch sizes), TENGraD still achieves the highest performance score. 

All hyperparameters used in all the experiments can be found in our code repository. 

\subsection{Batch Size Tuning}

For the batch size model, we tuned the batch size for five million environment steps with values (25 20 15 10 5) except for TENGraD, because of limited GPU memory, which we searched over values (10 8 6 4). EKFAC's performance on HalfCheetah-v2 (which it was tuned on) decreased with the tuned batch size so we used the baseline batch size when applying the natural critic optimization.

\subsection{Natural Critic Optimization}

For the natural critic optimization model, we applied a grid search over the following hyperparameters for five million environment steps: natural gradient approximation with values (Diagonal HF KFAC EKFAC TENGraD), critic damping parameter with values (1e-1 1e-2), whether to use a backtracking line search or not based on \cite{trpo}, when not using line search we use step-size clipping used in \cite{acktr} and tune the maximum learning rate with values (1e-1 1e-2).

\section{Model Architectures}

The policy and critic models consisted of three linear layers with 64 hidden units each using the tanh activation function, except for the final output. Following standard practice, all layers were initialized with orthogonal initialization.

\section{Code}

For fisher approximation implementations we implemented the diagonal ourselves, we used \cite{pytroch_trpo} as a reference when implementing HF. we used the code provided by EKFAC \cite{ekfac} for EKFAC and KFAC, and we used the author's provided code for the implementation of TENGraD. 

\begin{table*}[ht]
    \centering

    \begin{tabular}{lllll}
    \toprule
    {Environment} &  Performance &     Stability & Sample Efficiency &       Speed \\
    \midrule
    HalfCheetah-v2 &  1371 (+36\%) &   -257 (+51\%) &      -2250 (-22\%) &  -22 (+18\%) \\
    Ant-v2         &   \textbf{668 (+57\%)} &  \textbf{-100 (+100\%)} &      -1575 (-38\%) &  -43 (+17\%) \\
    Hopper-v2      &  1500 (+29\%) &   -330 (+10\%) &       \textbf{-1125 (+5\%)} &  -27 (+17\%) \\
    Humanoid-v2    &    749 (+3\%) &    -42 (+42\%) &       -4775 (-9\%) &  \textbf{-50 (+19\%)} \\
    Walker2d-v2    &   1802 (+0\%) &   -210 (+40\%) &      -1275 (-89\%) &  -28 (+18\%) \\
    Reacher-v2     &    -12 (+8\%) &  -1912 (+48\%) &       -775 (-67\%) &  -21 (+18\%) \\
    Swimmer-v2     &     36 (-2\%) &     -4 (+18\%) &      -4025 (-19\%) &  -22 (+18\%) \\
    \bottomrule
    \end{tabular}
    \caption{Performance of the \textit{Diagonal} approximation across seven different environments with a tuned batch size. The percentage improvement compared to the baseline is in brackets. The largest improvement for each metric is in bold.}
    \label{tab:diag_batch_tuned}

\end{table*}

\begin{table*}[ht]
    \centering

    \begin{tabular}{lllll}
    \toprule
    {Environment} &  Performance &     Stability & Sample Efficiency &       Speed \\
    \midrule
    HalfCheetah-v2 &  1338 (+17\%) &  -171 (-314\%) &      \textbf{-1900 (+35\%)} &   -28 (-5\%) \\
    Ant-v2         &  1123 (+18\%) &    -92 (-85\%) &      -2630 (+17\%) &   -54 (-3\%) \\
    Hopper-v2      &   2358 (+3\%) &   -340 (-14\%) &       -510 (+17\%) &   -35 (-4\%) \\
    Humanoid-v2    &   \textbf{991 (+21\%)} &    -47 (-42\%) &      -3680 (+16\%) &   -63 (-3\%) \\
    Walker2d-v2    &   3065 (+8\%) &    -298 (+7\%) &        -690 (+6\%) &   -37 (-5\%) \\
    Reacher-v2     &     -5 (+1\%) &     \textbf{-0 (+24\%)} &       -360 (+25\%) &   -28 (-7\%) \\
    Swimmer-v2     &     41 (-1\%) &     -5 (-49\%) &       -1200 (+5\%) &  -29 (-10\%) \\
    \bottomrule
    \end{tabular}
    \caption{Performance of the \textit{HF} approximation across seven different environments with a tuned batch size. The percentage improvement compared to the baseline is in brackets. The largest improvement for each metric is in bold.}
    \label{tab:hf_batch_tuned}

\end{table*}


\begin{table*}[ht]
    \centering

    \begin{tabular}{lllll}
    \toprule
    {Environment} &  Performance &    Stability & Sample Efficiency &      Speed \\
    \midrule
    HalfCheetah-v2 &   2200 (-3\%) &  -460 (+11\%) &       -430 (+25\%) &  -28 (-5\%) \\
    Ant-v2         &  \textbf{1725 (+10\%)} &  -346 (-18\%) &       -400 (+33\%) &  -54 (-4\%) \\
    Hopper-v2      &  1097 (-43\%) &   -457 (+5\%) &               NaN &  \textbf{-34 (+4\%)} \\
    Humanoid-v2    &    788 (+8\%) &  -80 (-148\%) &      -3940 (+17\%) &  -65 (-4\%) \\
    Walker2d-v2    &   1337 (-7\%) &  -519 (-66\%) &       -400 (+33\%) &  -36 (-3\%) \\
    Reacher-v2     &    -8 (-14\%) &   \textbf{ -1 (+17\%)} &       \textbf{-370 (+37\%)} &  -27 (-4\%) \\
    Swimmer-v2     &     41 (-1\%) &     -7 (+1\%) &     -2430 (-125\%) &  -29 (-8\%) \\
    \bottomrule
    \end{tabular}
    \caption{Performance of the \textit{EKFAC} approximation across seven different environments with a tuned batch size. The percentage improvement compared to the baseline is in brackets. The largest improvement for each metric is in bold.}
    \label{tab:ekfac_batch_tuned}

\end{table*}

\begin{table*}[ht]
    \centering

    \begin{tabular}{lllll}
    \toprule
    {Environment} &  Performance &    Stability & Sample Efficiency &      Speed \\
    \midrule
    HalfCheetah-v2 &   1943 (+8\%) &   -371 (-6\%) &      -1220 (+25\%) &  -42 (-1\%) \\
    Ant-v2         &  1998 (+10\%) &  -236 (-28\%) &      -1356 (+14\%) &  -69 (-1\%) \\
    Hopper-v2      &   1870 (+7\%) &  -318 (-40\%) &       -340 (+11\%) &  -49 (-1\%) \\
    Humanoid-v2    &   \textbf{972 (+13\%)} &   \textbf{ -49 (+6\%)} &      -3728 (+11\%) &  -79 (-2\%) \\
    Walker2d-v2    &   2788 (+2\%) &  -576 (-27\%) &       -496 (+27\%) &  -51 (-0\%) \\
    Reacher-v2     &    -6 (-14\%) &    -1 (-63\%) &       -268 (+27\%) &  -42 (-0\%) \\
    Swimmer-v2     &     45 (-3\%) &    -6 (-24\%) &       \textbf{-452 (+64\%)} &  -43 (-1\%) \\
    \bottomrule
    \end{tabular}
    \caption{Performance of the \textit{TENGraD} approximation across seven different environments with a tuned batch size. The percentage improvement compared to the baseline is in brackets. The largest improvement for each metric is in bold.}
    \label{tab:tengrad_batch_tuned}

\end{table*}

\end{document}